\documentclass[submission,copyright,creativecommons]{eptcs}
\providecommand{\event}{ICLP 2021} 
\usepackage{breakurl}             
\usepackage{underscore}           

\newcommand {\sx} {\langle}
\newcommand {\dx} {\rangle}

\newcommand {\emme} {\mathcal{M}}

\newcommand {\tc} {\mid}
\newcommand {\vuoto} {\emptyset}

\newcommand{\tip}{{\bf T}}

\newcommand{\alct}{\mathcal{ALC}+\tip}

\newcommand{\elb}{\mathcal{EL}^{\bot}}
\newcommand{\el}{\mathcal{EL}}
\newcommand{\elpb}{{\mathcal{EL}}^{+}_{\bot}}

\newcommand{\be}{\begin{enumerate}}
\newcommand{\ee}{\end{enumerate}}

\newcommand{\hide}[1]{}

\def \cases{\left \{\begin{array}{l}}
\def \endcases{\end{array}\right .}

\newcommand {\bes} {\begin{description}}
\newcommand{\ens} {\end{description}}

\newcommand {\beq} {\begin{quote}}
\newcommand {\enq} {\end{quote}}
\newcommand {\bit} {\begin{itemize}}
\newcommand {\enit} {\end{itemize}}

\usepackage{times}
\usepackage{helvet}
\usepackage{courier}
\usepackage{times}
\usepackage{graphicx}
\usepackage{latexsym}
\usepackage{graphicx}
\usepackage{color}
\usepackage{amssymb}
\usepackage{colortbl}
\usepackage{amsmath}
\usepackage{microtype}
\usepackage{amsmath}
\usepackage{tikz}

\newtheorem{definition}{Definition}

\title{Weighted Conditional $\elb$ Knowledge Bases \\
with Integer Weights:  an ASP Approach}
\author{Laura Giordano \qquad\qquad Daniele Theseider Dupr{\'{e}} 
\institute{DISIT - Universit\`a del Piemonte Orientale, 
 Alessandria, Italy }
 \email{\{laura.giordano,dtd\}@uniupo.it}
}
\def\titlerunning{Weighted Conditional $\elb$ KBs with Integer Weights: an ASP Approach}
\def\authorrunning{L. Giordano \& D. Theseider Dupr{\'{e}}}

\begin{document}
\bibliographystyle{eptcs}

\maketitle 

\begin{abstract}

Weighted knowledge bases for description logics with typicality  have been recently considered under a ``concept-wise'' multipreference semantics (in both the two-valued and fuzzy case), as the basis of a logical semantics of  Multilayer Perceptrons.
In this paper we consider weighted conditional $\elb$ knowledge bases  in the two-valued case, and  
exploit  ASP and {\em asprin} for encoding concept-wise multipreference entailment for weighted KBs with integer weights.

\end{abstract}

\section{Introduction}

Preferential approaches to common sense reasoning \cite{KrausLehmannMagidor:90,Pearl90,whatdoes,BenferhatIJCAI93,Kern-Isberner01},
have been extended to description logics (DLs), to deal with inheritance with exceptions in ontologies,
by allowing for non-strict forms of inclusions,
called {\em typicality or defeasible inclusions},
with different preferential semantics  \cite{lpar2007,sudafricaniKR} 
and closure constructions \cite{casinistraccia2010,CasiniDL2013,AIJ15,Pensel18,CasiniStracciaM19}.

In recent work, a concept-wise multipreference semantics  \cite{TPLP2020} has been proposed as a semantics of ranked $\elpb$ knowledge bases, 
 i.e. knowledge bases in which defeasible or typicality inclusions of the form $\tip(C) \sqsubseteq D$ (meaning ``the typical $C$'s are $D$'s" or ``normally $C$'s are $D$'s") are given a rank, a natural number, representing 
their strength, where $\tip$ is a typicality operator \cite{lpar2007} that singles out the typical instances of concept $C$. 
The concept-wise multipreference semantics takes into account preferences with respect to different concepts, and integrates them into a single global preference relation,  which is needed for the evaluation of general defeasible inclusions.
Answer Set Programming (ASP) and, in particular, {\em asprin} \cite{BrewkaAAAI15}, has been exploited to achieve defeasible reasoning under the multipreference approach for the lightweight description logic $\elpb$ (a logic of the $\el$ family \cite{rifel}).

In \cite{JELIA2021}, the concept-wise multipreference semantics has been adapted to deal with weighted knowledge bases, in which each typicality inclusion is associated to a weight, a real (positive or negative) number, representing its plausibility or implausibility. 
The multipreference semantics has been exploited to provide a preferential interpretation of Multilayer Perceptrons (MLPs), by representing the input-output behavior of the network. It has been further extended to the fuzzy case, and the fuzzy multipreference interpretation built from the network, for a given set of input stimuli, have been proven to be a model of the neural network, in the logical sense. To this purpose, the deep network has been interpreted as a weighted conditional knowledge base, i.e., a set of weighted defeasible inclusions,
by regarding synaptic connections as weighted defeasible inclusions.

While decidability of the fuzzy multipreference entailment is an open problem, in this paper we consider the two-valued case and extend the ASP approach considered for ranked KBs \cite{TPLP2020} to deal with weighted conditional KBs, for integer weights. 
Differently from \cite{TPLP2020}, the semantic closure construction used for weighted knowledge bases does not exploit a lexicographic order, as in  Lehmann lexicographic closure \cite{whatdoes} and in Brewka's framework of basic preference descriptions  \cite{Brewka04}, but a construction more related to Kern-Isberner's c-representations  \cite{Kern-Isberner01,Kern-Isberner2014}. 
We develop an ASP encoding of the  concept-wise multipreference semantics for weighted $\elb$ knowledge bases,  
and exploit {\em asprin} to achieve defeasible reasoning for weighted $\elb$ knowledge bases with integer weights.
This is a first step towards the definition of multi-valued  approximations of 
the fuzzy multipreference semantics for weighted knowledge bases, which may be of interest from the standpoint of explainable AI \cite{Adadi18,Guidotti2019}. 

\section{The description logic $\elb$}

We consider the description logic $\elb$ of the ${\cal EL}$ family  \cite{rifel}. 
Let ${N_C}$ be a set of concept names, ${N_R}$ a set of role names
  and ${N_I}$ a set of individual names.  
The set  of $\elb$ \emph{concepts} can be
defined as follows: 
  $C \ \ := A \tc \top \tc \bot  \tc C \sqcap C \tc \exists r.C \tc \{a\}$
where $a \in N_I$, $A \in N_C$ and $r \in N_R$. $\{a\}$ is a nominal, a concept containing a single element.
Observe that union, complement and universal restriction are not $\elb$ constructs.
A knowledge base (KB) $K$ is a pair $({\cal T}, {\cal A})$, where ${\cal T}$ is a TBox and
${\cal A}$ is an ABox.
The TBox ${\cal T}$ is  a set of {\em concept inclusions} (or subsumptions) of the form $C \sqsubseteq D$, where $C,D$ are concepts.
The  ABox ${\cal A}$ is  a set of assertions of the form $C(a)$ 
and $r(a,b)$ where $C$ is a  concept, $r \in N_R$, and $a, b \in N_I$.

An {\em interpretation} for $\elb$ is a pair $I=\langle \Delta, \cdot^I \rangle$ where:
$\Delta$ is a non-empty domain---a set whose elements are denoted by $x, y, z, \dots$---and 
$\cdot^I$ is an extension function that maps each
concept name $C\in N_C$ to a set $C^I \subseteq  \Delta$, each role name $r \in N_R$
to  a binary relation $r^I \subseteq  \Delta \times  \Delta$,
and each individual name $a\in N_I$ to an element $a^I \in  \Delta$.
It is extended to complex concepts  as follows:
$\top^I=\Delta$, $\bot^I=\vuoto$,  $\{a\}^I=\{a^I\}$, 
$(C \sqcap D)^I =C^I \cap D^I$  and 
$(\exists r.C)^I =\{x \in \Delta \tc \exists y.(x,y) \in r^I \ \mbox{and} \ y \in C^I\}.$	

The notion of satisfiability of a KB  in an interpretation 
is defined as usual:

\begin{definition}[Satisfiability and Entailment] \label{satisfiability}
Given an $\elb$ interpretation $I=\langle \Delta, \cdot^I \rangle$: 

	- $I$  satisfies an inclusion $C \sqsubseteq D$ if   $C^I \subseteq D^I$;

	-  $I$ satisfies an assertion $C(a)$ if $a^I \in C^I$ and an assertion $r(a,b)$ if $(a^I,b^I) \in r^I$.

\noindent
 Given  a KB $K=({\cal T}, {\cal A})$, 
 an interpretation $I$  satisfies ${\cal T}$ (resp. ${\cal A}$) if $I$ satisfies all  inclusions in ${\cal T}$ (resp. all assertions in ${\cal A}$);
 $I$ is a \emph{model} of $K$ if $I$ satisfies ${\cal T}$ and ${\cal A}$.

 A subsumption $F= C \sqsubseteq D$ (resp., an assertion $C(a)$, $R(a,b)$),   {is entailed by $K$}, written $K \models F$, if for all models $I=$$\sx \Delta,  \cdot^I\dx$ of $K$, $I$ satisfies $F$.
\end{definition}

\section{Weighted knowledge bases and the multipreference semantics} \label{sec:WeightedBKs}

Let ${\cal C}= \{C_1, \ldots, C_k\}$ be a set of distinguished $\elb$ concepts, the concepts for which defeasible inclusions are defined. 
A weighted TBox ${\cal T}_{C_i}$ is defined for each distinguished concept $C_i \in {\cal C}$ as a set of defeasible inclusions of the form $\tip(C_i) \sqsubseteq D$ with a weight.
A {\em weighted $\elb$ knowledge base $K$ over ${\cal C}$} is a tuple $\langle  {\cal T}_{strict}, {\cal T}_{C_1}, \ldots, {\cal T}_{C_k}, {\cal A}  \rangle$, 
where ${\cal T}_{strict}$ is a set of strict concept inclusions, ${\cal A}$ is an ABox and, for each $C_i \in {\cal C}$,  ${\cal T}_{C_i}$ is 
a set of weighted defeasible inclusions,
$\{(d^i_h,w^i_h)\}$, where  each  $d^i_h$ is a typicality inclusion of the form $\tip(C_i) \sqsubseteq D_{i,h}$,  having weight $w^i_h$, a real number.

Consider, for instance, the weighted knowledge base $K =\langle {\cal T}_{strict},  {\cal T}_{Emp}, {\cal T}_{Student},$ $ {\cal A} \rangle$, over the set of distinguished concepts ${\cal C}=\{\mathit{Emp, Student}\}$, with empty ABox,
and with $ {\cal T}_{strict}$ containing the set of strict inclusions:

 \ \ \ \ \ \ \ \ \ \ \ \ \ \ \   $\mathit{Emp  \sqsubseteq  Adult}$ \ \ \ \ \ \ \ \ \ \ \ \ \ \ \   $\mathit{Adult  \sqsubseteq  \exists has\_SSN. \top}$  
 \ \ \ \ \ \ \ \ \ \ \   $\mathit{PhdStudent  \sqsubseteq  Student}$

\noindent
${\cal T}_{Emp} $ 
contains the following weighted defeasible inclusions: 
  
$(d_1)$ $\mathit{\tip(Emp) \sqsubseteq Young}$, \ \ - 50  \ \ \ \ \ \ \ \ \ \ \ \ \ \ \ \ \ \
$(d_2)$ $\mathit{\tip(Emp) \sqsubseteq \exists has\_boss.Emp}$, \ \ 100

$(d_3)$ $\mathit{\tip(Emp) \sqsubseteq  \exists has\_classes.\top}$, \ \ -70;

\noindent
 ${\cal T}_{Student}$ contains the defeasible inclusions:

$(d_4)$ $\mathit{\tip(Student) \sqsubseteq Young}$, \ \ 90
 \ \ \ \ \ \ \ \ \ \ \ \ \ $(d_5)$ $\mathit{\tip(Student) \sqsubseteq  \exists has\_classes.\top}$, \ \ 80

$(d_6)$ $\mathit{\tip(Student) \sqsubseteq  \exists hasScholarship.\top}$, \ \  -30

\noindent
The meaning is that, while an employee normally has a boss, he is not likely to be young or have classes. Furthermore, between the two defeasible inclusions $(d_1)$ and $(d_3)$, the second one is considered to be less plausible than the first one. 
Given two employees Tom and Bob such that  Tom is not young, has no boss and has classes, while  Bob is not young, has a boss and has no classes, considering the weights above, we will regard Bob as being more typical than Tom as an employee.  Note that negative weights represent penalties that could not be expressed by positive ranks in a ranked $\elb$ knowledge base.

The semantics of a weighted knowledge base has been defined \cite{JELIA2021} based on a concept-wise preference semantics, a semantics first exploited for ranked $\elb$ knowledge bases  \cite{TPLP2020}.
For each concept $C_i \in {\cal C}$, a preference relation $<_{C_i}$ describes the preference among domain elements with respect to $C_i$. 
Each $<_{C_i}$ has the properties of preferences in KLM-style ranked interpretations \cite{whatdoes}, i.e.,  it is a modular and well-founded strict partial order. 
More precisely, $<_{C_i}$ is {\em well-founded} 
if, for all $S \subseteq \Delta$, if $S\neq \emptyset$, then $min_{<_{C_i}}(S)\neq \emptyset$;
    $<_{C_i}$ is {\em modular} if,
for all $x,y,z \in \Delta$, $x <_{C_j} y$ implies ($x <_{C_j} z$ or $z <_{C_j} y$).

To define a concept-wise semantics for weighted $\elb$ KBs, let us recall the notion of multipreference interpretation \cite{TPLP2020}:
an $\elb$ interpretation is extended with a collection of preferences.
  \begin{definition}[Multipreference interpretation]\label{defi:multipreference}  
A {\em multipreference $\elb$ interpretation}  is a tuple 
$\emme= \langle \Delta, <_{C_1}, \ldots, <_{C_k}, \cdot^I \rangle$, where:

(a) $\Delta$ is a 
domain, and $\cdot^I$ an interpretation function, as in $\elb$ interpretations;
 
(b) for each $C_i \in {\cal C}$,  $<_{C_i}$ is an irreflexive, transitive, well-founded and modular relation over $\Delta$.

\end{definition}
A preference relation $<_{C_i}$ determines the relative typicality of domain individuals with respect to concept $C_i$.
For instance,  Tom may be more typical than Bob as a student ($\mathit{tom <_\mathit{Student} bob}$), but more exceptional than Tom as an employee ( $\mathit{bob <_\mathit{Emp} tom}$). 
The minimal $C_i$-elements with respect to $<_{C_i}$ 
are 
taken as the most typical  $C_i$-elements.

For a multipreference interpretation $\emme$ to satisfy a weighted knowledge base $K= \langle  {\cal T}_{strict}, {\cal T}_{C_1}, \ldots,$ $ {\cal T}_{C_k}, {\cal A}  \rangle$, we require $ \langle \Delta, \cdot^I \rangle$ to satisfy the axioms in ${\cal T}_{strict}$ and  ${\cal A}$, as usual in $\elb$, and each preference relation $<_{C_i}$ to be constructed from ${\cal T}_{C_i}$, through a {\em semantic closure construction}, similar in spirit to the lexicographic closure \cite{Lehmann95}, but more similar to to c-representation \cite{Kern-Isberner01,Kern-Isberner2014}.
The sum of the weights of the defeasible inclusions for $C_i$ satisfied by each domain element $x \in \Delta$ is considered;  higher preference wrt $<_{C_i}$ is given to the domain elements whose associated sum (wrt $C_i$) is higher \cite{JELIA2021}.

As $\elb$ has the finite model property \cite{rifel},  we can restrict to interpretations with a finite domain $\Delta$.
We say that {\em$ x \in \Delta$  satisfies $\tip(C_i) \sqsubseteq D$ in $I$}, if  $x   \not \in C_i^I$ or $x \in D^I$ (otherwise $x$ {\em violates} $\tip(C_i) \sqsubseteq D$ in $I$).
Given an $\elb$ interpretation $I=\langle \Delta, \cdot^I \rangle$ and a domain element $x \in \Delta$,  the {\em weight $W_i(x)$ of $x$ wrt $C_i$ in $I$}  
is defined considering the inclusions $(\tip(C_i) \sqsubseteq D_{i,h} \; , w^i_h)\in {\cal T}_{C_i}$, as follows:  
\vspace{-0.2cm}
\begin{align}\label{weight}
	W_i(x)  & = \left\{\begin{array}{ll}
						\sum_{{h: x \in D_{i,h}^I}} w_h^i & \mbox{ \ \ \ \  if }  x \in C_i^I \\
						- \infty &  \mbox{ \ \ \ \  otherwise }  
					\end{array}\right.
\end{align}
where $-\infty$ is added at the bottom of all real values.
Informally, given an interpretation $I$, for  $x \in C_i^I$, the weight $W_i(x)$ of $x$ wrt $C_i$ is the sum of the weights of all defeasible inclusions for $C_i$ satisfied by $x$ in $I$. The more plausible are the satisfied inclusions, the higher is the weight of $x$. 
The lowest weight, $-\infty$, is given to all domain elements which are not instances of $C_i$.

Based on this notion of weight of a domain element wrt a concept, one can construct a preference relation $<_{C_i}$ from a given $\elb$ interpretation $I$. A domain element $x$ is preferred to element $y$ wrt $C_i$ 
if the sum of the weights of the defaults in  ${\cal T}_{C_i}$ satisfied by $x$ is higher than the sum of the weights of defaults in $ {\cal T}_{C_i}$ satisfied by $y$:
for $x,y \in \Delta$, 
\begin{align}  \label{pref}
x & <_{C_i}  y  \mbox{  \ \ iff \ \ }  W_i(x) > W_i(y)
\end{align}
 Note that $<_{C_j}$ a strict modular partial order, and all $C_i$-elements are preferred wrt  $<_{C_i}$ to the domain elements which are not instances of $C_i$.
The higher is the weight of an element wrt $C_i$ the more preferred is the element.
In the example, $W_i(bob)=30 > W_i(tom)=-70$  (for $C_i= \mathit{Emp}$) and, hence, $\mathit{bob <_{Emp} tom}$, i.e., Bob is more typical than Tom as an employee.

Given the preferences $<_{C_i}$ for the distinguished concepts, we can interpret $\tip(C_i)$ as  the set of minimal $C_i$ elements w.r.t. preference $<_{C_i}$. 
To provide an interpretation of the typicality concept $\tip(C)$ for an arbitrary $C$ (such as, for instance, $\mathit{ \tip(Employee \sqcap Student)}$),
following  \cite{TPLP2020},
a notion of global preference $<$ is introduced by exploiting a modified {\em Pareto} combination of the preference relations $<_{C_1}, \ldots,<_{C_k}$,
which takes into  account the {\em specificity relation} $\succ$ among concepts, e.g., that concept $\mathit{PhDStudent}$ is more specific than concept  $\mathit{Student}$ ($\mathit{PhDStudent \succ Student}$), and its properties override the properties of $\mathit{Student}$, when conflicting). The global preference relation $<$ is defined from  $<_{C_1}, \ldots,<_{C_k}$ as follows: \\
$\mbox{\ \ \ \ \ \ \ \ \ \ \ \ \ \ \ \ \ \ \ \ \ \ \ \ \ }$ $x <y  \mbox{ iff \ \ }  (i) \  x <_{C_i} y, \mbox{ for some } C_i \in {\cal C}, \mbox{ and }$ \\
$\mbox{\ \ \ \ \ \ \ \ \ \ \ \ \ \ \ \ \ \ \ \ \ \ \ \ \ \ \ \ \ \  \ \ \ \ \ \ \ \ \ \ }$ $(ii)  \ \mbox{  for all } C_j\in {\cal C}, \;  x \leq_{C_j} y \mbox{ or }  \exists C_h (C_h \succ C_j  \mbox{ and } x <_{C_h} y )$.\\
We interpret $\tip(C)$, for  an arbitrary concept $C$, 
 as the set of minimal $C$-elements with respect to $<$ (i.e., $(\tip(C))^I= min_{<} (C^I)$).
This leads to the definition of a {\em concept-wise multipreference model} (cw$^m$-model) of a weighted knowledge base
$K=\langle  {\cal T}_{strict},$ $ {\cal T}_{C_1}, \ldots,$ $ {\cal T}_{C_k}, {\cal A}  \rangle$ over  ${\cal C}$,  as a 
a tuple ${\emme}=\langle \Delta,<_{C_1}, \ldots, <_{C_k}, <, \cdot^I \rangle$, where
$\langle \Delta, \cdot^I \rangle$  satisfies  
$ {\cal T}_{strict}$  and ${\cal A}$ in $\elb$;
$<_{C_j}$ is defined from  ${\cal T}_{C_j}$ and $I$, according to condition (\ref{pref}), 
 for all $j= 1, \ldots, k$; and $<$ is the global preference relation.

Based on the notion of cw$^m$-model of a KB, a notion of  {\em concept-wise entailment}  (or cw$^m$-entailment) can be defined in a natural way for weigthed KBs.
Let us restrict consideration 
 to (finite) {\em canonical } models, i.e., models which are large enough to contain 
all the relevant domain elements (see \cite{TPLP2020}).

\begin{definition}[cw$^m$-entailment  \cite{JELIA2021}] \label{cwm-entailment}
An inclusion $\tip(C) \sqsubseteq D$ is cw$^m$-entailed  
from a weighted knowledge base $K$ 
if 
it is satisfied in all canonical cw$^m$-models  
$\emme$ of $K$.
\end{definition}
As for ranked $\elpb$ knowledge bases \cite{TPLP2020}, it can be proven that this notion of cw$^m$-entailment for weigthed $\elb$ KBs satisfies the KLM postulates of a preferential consequence relation, as the global preference relation $<$ is a strict partial order.

\section{Encoding cw$^m$-entailment in ASP and {\em asprin} for integer weights}

From the computational point of view, in the case of knowledge bases integer weights, the notion of cw$^m$-entailment 
can be reformulated as a problem of computing preferred answer sets, as  done for ranked knowledge bases \cite{TPLP2020}.
It is possible to adapt the ASP encoding for ranked knowledge bases to deal with the extension of $\elb$ with typicality inclusions with integer weights.
The encoding has been developed by exploiting a fragment of Kr\"{o}tzsch's Datalog materialization calculus \cite{KrotzschJelia2010} to generate the answer sets (representing canonical cw$^m$-models), 
 and {\em asprin} \cite{BrewkaAAAI15}  to select preferred answer sets: {\em asprin} allows to do this according to preferences defined from a library, or with a preference program, as in this case.
 
In principle, verifying cw$^m$-entailment of a typicality subsumption $\tip(C) \sqsubseteq D$ from $K$,
would require considering all typical $C$-elements in all possible canonical cw$^m$-models of $K$, and checking whether they are all instances of $D$.
However, 
it is sufficient to consider,
among all the (finite) cw$^m$-models of $K$, the polynomial $\elb$ models that we can construct using the $\elb$ fragment of the materialization calculus by Kr\"{o}tzsch 
\cite{KrotzschJelia2010}, 
and a distinguished domain element $aux_C$ to represent a prototypical $C$-element. 
The preferred answer sets are those  maximizing the weight of typicality inclusions  satisfied by $aux_C$.
As in the materialization calculus auxiliary constants 
are used to deal with existential rules. 
Differently from \cite{TPLP2020}, we do not need other auxiliary predicates for the distinguished concepts $C_i$.

Following \cite{KrotzschJelia2010}, 
we assume that the knowledge base $K$ is in {\em normal form} \cite{rifel}, where a typicality inclusion  $\tip(B) \sqsubseteq C$ is in normal form when $B, C \in N_C$.  
Extending the results by Kr\"{o}tzsch \cite{KrotzschJelia2010}, it can be proven that, given a KB, a semantically equivalent KB in normal form (over an extended signature) can be computed in linear time  \cite{SROEL-FI2018}.

The base program $\Pi(K,C,D)$ for the (normalized) knowledge base $K$ and a typicality subsumption 
$\tip(C) \sqsubseteq D$ is composed of three parts, 
$\Pi(K,C,D)= \Pi_{K} \cup \Pi_{IR} \cup \Pi_{C,D}$.
$\Pi_{K}$ is the representation of $K$ in Datalog based on the materialization calculus \cite{KrotzschJelia2010}, 
where $\mathit{nom(a)}$, $\mathit{cls(A)}$, $\mathit{rol(R)}$ are used for
$\mathit{a \in N_I}$ , $\mathit{A \in N_C}$, $\mathit{R \in N_R}$, and, for example,
	$\mathit{subClass(a,C)}$ and $\mathit{subClass(A,C)}$ are used, respectively, for $C(a)$ and  $A \sqsubseteq C$.
Additionally, $\mathit{subTyp(C,D,W)} $ is introduced to describe the typical properties $ \mathit{T(C) \sqsubseteq D} $ of concept $C$ with their weight $W$, and is used for specifying preferences in {\em asprin}. 

$\Pi_{IR}$ contains the subset  of the
inference rules 
for instance checking from the materialization calculus \cite{KrotzschJelia2010}, those relevant for $\elb$, 
for example,	$\mathit{inst(x,z) \leftarrow}$ $\mathit{ subClass(y,z),inst(x,y)} $. For
$\bot$, we use an additional rule: $ \mathit{\leftarrow bot(z), inst(x,z)}$. 
Additionally, $\Pi_{IR}$ contains the version of the same rules for subclass checking  
(where $\mathit{inst\_sc(A,B,A)}$ represents $A  \sqsubseteq B$ \cite{KrotzschJelia2010}),
and a rule to define predicate  $\mathit{morespec(Ch,Cj)}$ describing the specificity relation  among concepts (meaning that concept $C_h$ is more specific than concept $C_j$).
$\Pi_{IR}$ also contains the rule:
$ \mathit{
\{inst(aux_C,D)\}\ \leftarrow cls(D)
	} $,
which generates alternative answer sets, corresponding to different interpretations of $\mathit{aux_C}$, which may be an instance or not of any concept $D$.  

Let us notice that the rules enforcing $\tip$-compliance 
in \cite{TPLP2020} are omitted. 
In fact, in a $\tip$-compliant interpretation all typical $C_i$ elements are assumed to satisfy {\em all} typicality inclusions for $C_i$ in ${\cal T}_{C_i}$, an assumption 
which cannot be taken in the presence of defeasible inclusions with negative weights. 

$\Pi_{C,D}$ contains (when needed) normalized axioms defining $C,D$ in $\tip(C) \sqsubseteq D$ in terms of other concepts 
(e.g., replacing $\mathit{\tip(Emp \sqcap Student) \sqsubseteq Young}$ with $\mathit{\tip(A) \sqsubseteq Young}$ and $\mathit{A \sqsubseteq Emp}$,  $\mathit{A \sqsubseteq Student}$ and $\mathit{Emp \sqcap Student \sqsubseteq A}$)
plus the facts
$ \mathit{auxtc(aux_C,C)}$, $ \mathit{nom(aux_C)}$, $ \mathit{inst(aux_C,C).}$

 Differently from  the ASP encoding in \cite{TPLP2020}, where the lexicographic strategy  \#  \cite{Brewka04} for ranked knowledge bases is used,
 the preference among individuals is now expressed as: $x <_{C_i} y$ iff $W_i(x) > W_i(y)$, 
 where $W_i(x)$ and $W_i(y)$ are 
 integer values.
 Given a query $\tip(C) \sqsubseteq D$, we have to verify that, in all canonical cw$^m$-models of the KB, all the typical $C$-elements are $D$-elements.
This verification is accomplished, by generating answer sets, corresponding to the cw$^m$-models of the KB, and by selecting the preferred ones, in which the distinguished element $aux_C$ represents a typical $C$-element.

An answer set $S$ is preferred to answer set $S'$ if, considering the typicality inclusions satisfied by $aux_C$ in $S$ and in $S'$ (and the relative weights $W_i(aux_C^S)$ and 
$W_i(aux_C^{S'})$ for each concept $C_i$), in the global preference relation, $aux_C^S < aux_C^{S'}$. 
The preference relations $ <_{C_i}$ can be suitably encoded  in the {\em aspirin} specification (based on the definition above), verifying which atoms $\mathit{inst(aux_C,B)}$ are in $S$ (resp., in $S'$) to determine the typicality inclusions satisfied by $aux_C^S$ and $aux_C^{S'}$,  
to establish whether  $aux_C^S <_{C_i} aux_C^{S'}$.
The (global) preference relation between answer sets can be defined by combining the preferences $<_{C_i}$.
A query $\tip(C) \sqsubseteq D$ is entailed by the knowledge base if, in all the preferred answer sets, $aux_C$ is an instance of concept $D$.
It can be proven that this corresponds to verifying that  $D$ is satisfied in all minimal $C$-elements in all canonical cw$^m$-models of the knowledge base.

Based on the reformulation 
of cw$^m$-entailment as a problem of computing preferred answer sets,  
deciding cw$^m$-entailment can be proven to be in $\Pi^p_2$ for weighted knowledge bases with integer weights, as for ranked KBs. 
The proof of the result is similar to the proof of Proposition 7 in the online Appendix  of \cite{TPLP2020},
although here we use a different notion of preference $<_{C_i}$, 
a different encoding, and do not consider $\tip$-compliant interpretations.
This result also extends to weighted knowledge bases with real valued weights, although our implementation in {\em asprin} does not deal with real valued weights.

\section{Conclusions}

In this paper we have described an ASP approach for reasoning in a defeasible extension of the description logic $\elb$ with weighted typicality inclusions. We have considered the case with integer weights, for which an encoding of the concept-wise multipreference entailment can be defined using ASP and {\em asprin} \cite{BrewkaAAAI15}. 
We have further considered 
a concept-wise semantics with multiple preferences, in which any distinguished concept $C_i$ has its own set ${\cal T}_{C_i}$ of (weighted) typicality inclusions, and preference $<_{C_i}$.

Related semantics with multiple preferences have been proposed, starting from Brewka's framework of basic preference descriptions  \cite{Brewka04}, 
in system ARS, as a refinement of System Z by Kern-Isberner and Ritterskamp \cite{IsbernerRitterskamp2010};
in an extension of $\alct$ 
by Gil \cite{fernandez-gil};
in a refinement of rational closure by Gliozzi \cite{GliozziAIIA2016}; 
by associating multiple  preferences to roles by Britz and Varzinczak \cite{Britz2018}; in ranked $\el$ knowledge bases  
in \cite{TPLP2020}; and
in the first-order logic setting by Delgrande and Rantsaudis 
\cite{Delgrande2020}.

The development of proof methods for multipreference entailment is motivated by the relationship between weighted conditionals under the (fuzzy) concept-wise multipreference semantics and  multilayer perceptrons.  
Undecidability results for fuzzy description logics with general inclusion axioms 
\cite{CeramiStraccia2011,BorgwardtPenaloza12} 
motivate the investigation of multi-valued approximations of fuzzy multipreference entailment. 
This is a first step towards the definition of proof methods for multi-valued extensions of our concept-wise 
preferential semantics based on a notion of faithful interpretations \cite{ECSQARU2021}. 
Other possible extensions concern the definition of multiple typicality operators, based on the combination of selected concepts, and a temporal extension to capture the transient behavior of Multilayer Perceptrons.

\medskip
{\bf Acknowledgement:} We thank the referees for their helpful suggestions. This research is partially supported by INDAM-GNCS Project 2020.

\vspace{-0.2cm}


\end{document}